\documentclass[11pt,a4paper]{article}

\usepackage[hyperref]{emnlp2020}
\usepackage{times, amsmath, amssymb, array}
\usepackage[final]{graphicx}

\usepackage{caption}
\usepackage{subcaption}

\usepackage{latexsym}

\usepackage{multirow, textcomp, subcaption}
\usepackage{microtype}
\aclfinalcopy

\usepackage{graphicx}

\newcommand{\ignore}[1]{}

\title{To BERT or Not to BERT: Comparing Task-specific and Task-agnostic Semi-Supervised Approaches for Sequence Tagging}

\author{Kasturi Bhattacharjee ~~ Miguel Ballesteros ~~  \textbf{Rishita Anubhai} \\ \textbf{Smaranda Muresan}\thanks{~~Smaranda Muresan is an Amazon Scholar and a Research Scientist at the Data Science Institute, Columbia University} ~~ \textbf{Jie Ma}~~ \textbf{Faisal Ladhak} ~~ \textbf{Yaser Al-Onaizan} \\
    Amazon AI \\
    {\tt \footnotesize{\{kastb, ballemig, ranubhai, smaranm, jieman, faisall, onaizan\}@amazon.com}}}

\date{}

\begin{document}
\maketitle
\begin{abstract}
  Leveraging large amounts of unlabeled data using  Transformer-like architectures, like BERT, has gained popularity in recent times owing to their effectiveness in learning general representations that can then be further fine-tuned for downstream tasks to much success. However, training these models can be costly both from an economic and environmental standpoint. In this work, we investigate how to effectively use unlabeled data: by exploring the task-specific semi-supervised approach, Cross-View Training (CVT) and comparing it with task-agnostic BERT in multiple settings that include domain and task relevant English data. CVT uses a much lighter model architecture and we show that it achieves similar performance to BERT on a set of sequence tagging tasks, with lesser financial and environmental impact.
\end{abstract}

\section{Introduction} \label{sec: Intro}
Exploiting unlabeled data to improve performance has become the foundation for many natural language processing tasks. The question we investigate in this paper is how to effectively use unlabeled data: in a task-agnostic or a task-specific way. An example of the former is training models on  language model (LM) like objectives on a large unlabeled corpus to learn general representations, as in ELMo (Embeddings from Language Models) \cite{Peters:2018} and BERT (Bidirectional Encoder Representations from Transformers) \cite{devlin2019bert}. These are then reused in supervised training on a downstream task. These pre-trained models, particularly the ones based on the Transformer architecture ~\cite{vaswani2017attention}\footnote{Not only BERT, but other models like RoBERTa \cite{Liu2019RoBERTaAR} and BART \cite{lewis2019bart}} have achieved state-of-the-art results in a variety of NLP tasks, but come at a great cost financially and environmentally \cite{strubell-etal-2019-energy, schwartz2019green}. 

In contrast, Cross-View Training (CVT) \cite{clark2018semi} is a semi-supervised approach that uses unlabeled data in a \emph{task-specific} manner, rather than trying to learn general representations that can be used for many downstream tasks. Inspired by self-learning \cite{mcclosky2006effective, yarowsky1995unsupervised} and multi-view learning \cite{blum1998combining, xu2013survey}, the key idea is that the primary prediction module, which has an unrestricted view of the data, trains on the task using labeled examples, and makes task-specific predictions on unlabeled data. The auxiliary modules, with restricted views of the unlabeled data, attempt to replicate the primary module predictions. This helps to learn better representations for the task.

We present an experimental study that investigates different task-agnostic and task-specific approaches to use unsupervised data and evaluates them in terms of performance as well as financial and environmental impact. On the one hand, we use BERT in three different settings: 1) standard BERT setup in which BERT pretrained on a generic corpus is fine-tuned on a supervised task; 2) pre-training BERT on domain and/or task relevant unlabeled data and fine-tuning on a supervised task (Pretrained BERT); and 3) continued pretraining of BERT on domain and/or task relevant unlabeled data followed by fine-tuning on a supervised task (Adaptively Pretrained BERT) \cite{Gururangan2020DontSP}. On the other hand, we use CVT based on a much lighter architecture (CNN-BiLSTM) which uses domain and/or task relevant unlabeled data in a \emph{task-specific} manner. We experiment on several tasks framed as a sequence labeling problem: opinion target expression detection, named entity recognition and slot-labeling. We find that the CVT-based approach using less unlabeled data achieves similar performance with BERT-based models, while being superior in terms of financial and environmental cost as well.

\section{Background, Tasks and Datasets} \label{sec:taskdata} 
\begin{table}
\scalebox{0.8}{
    	\begin{tabular}{|>{\centering\arraybackslash}p{12mm}|>{\centering\arraybackslash}p{38mm}|>{\centering\arraybackslash}p{27mm}|} 
    \hline
	\textbf{Task} & \textbf{Labeled Data} & \textbf{Unlabeled Data} \\ \hline
         \multirow{2}*{OTE} & SE16-R Train: 2000; Test: 676 & Yelp-R: $\sim$32.5M \\ \cline{2-3}
          					& SE14-L Train: 3045; Test: 800 & Amazon-E: $\sim$95M \\ \hline
        
        \multirow{2}*{NER} & CONLL-2003 Train: 14987; Test: 3684 & \multirow{2}*{CNN-DM: $\sim$4M} \\ \cline{2-2}
        & CONLL-2012 Train: 59924; Test: 8262 & \\ \hline
        
        Slot-labeling & MIT-Movies Train: 9775; Test: 2443 & IMDb: $\sim $ 271K\\ \hline
    \end{tabular}
    }
    \caption{Number of sentences in unlabeled data and default train-test splits of the labeled datasets, for the various tasks.}
    \label{tab:labeled_data_stats}
\end{table}

Before presenting the models and their training setups, we discuss the relevant literature and introduce the tasks and datasets used for our experiments. 
We focus on three tasks:  opinion target expression (OTE) detection; named entity recognition (NER), and slot-labeling, each of which can be modeled as a sequence tagging problem \cite{xu-etal-2018-double, liu-etal-2019-towards, slotlabelbaseline}. The IOB sequence tagging scheme \cite{ramshaw1999text} is used for each of these tasks. 

\vspace{1mm}
\noindent\textbf{Related Work.}
The usefulness of continued training of large transformer-based models on domain/task-related unlabeled data has been shown recently \cite{Gururangan2020DontSP, alex2019adapt, xu-etal-2019-bert}, with a varied use of terminology for the process. \newcite{xu-etal-2019-bert} and \newcite{alex2019adapt} show gains of further tuning BERT using in-domain unlabeled data and refer to this as Post-training, and LM finetuning, respectively. More recently, \newcite{Gururangan2020DontSP} use the term Domain-Adaptive Pretraining and show benefits over RoBERTa \cite{Liu2019RoBERTaAR}. There have also been efforts to reduce model sizes for BERT, such as DistilBERT \cite{Sanh2019DistilBERTAD}, although these come at significant losses in performance.

\vspace{1mm}

\begin{figure}[tbp]
\centering
\begin{subfigure}[b]{0.4\textwidth}
	\includegraphics[width=1\linewidth]{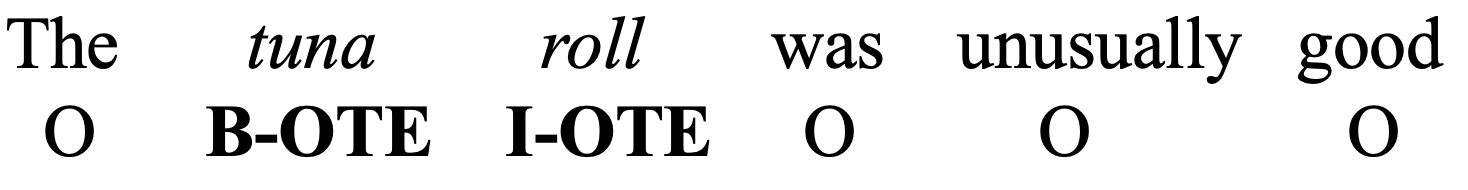}
	\caption{OTE detection example}
	\label{fig: seq example1}
\end{subfigure}

\hfill
\vspace{2mm}

\begin{subfigure}[b]{0.4\textwidth}
	\includegraphics[width=1\linewidth]{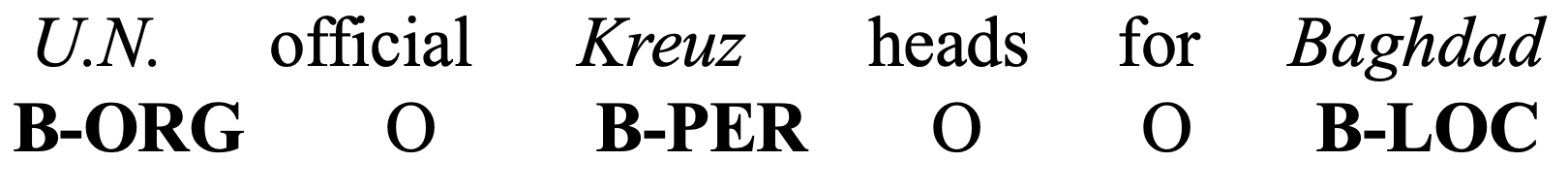}
	\caption{NER example}
	\label{fig: seq example2}
\end{subfigure}

\hfill
\vspace{2mm}

\begin{subfigure}[b]{0.4\textwidth}
	\includegraphics[width=1\linewidth]{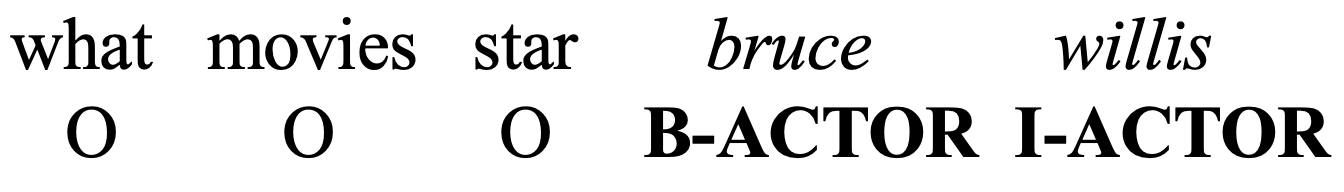}
	\caption{Slot labeling example}
	\label{fig: seq example3}
\end{subfigure}

\hfill

\caption{Examples illustrating each sequence tagging task studied here.}
\label{fig: seqlabel_examples}
\end{figure}

\paragraph{Opinion Target Expression (OTE) Detection:} An integral component of fine-grained sentiment analysis is the ability to identify segments of text towards which opinions are expressed. These segments are referred to as Opinion Target Expressions or OTEs. An example of this task is provided in Figure \ref{fig: seqlabel_examples}. 
The commonly used \emph{labeled} datasets for Opinion Target Expression (OTE) detection are those released as part of SemEval Aspect-based Sentiment shared tasks: SemEval-2014 Laptops \cite{pontiki-etal-2014-semeval} (\textbf{SE14-L}) and SemEval-2016 Restaurants \cite{pontiki2016semeval} (\textbf{SE16-R}). These consist of reviews from the laptop and restaurant domains, respectively, with OTEs annotated for each sentence of a review. We use the provided train-test splits but further split the training data randomly into 90\% training and 10\% validation sets. As \emph{unlabeled} data that is similar to the domain and task, we extract restaurant reviews from the Yelp\footnote{\url{https://www.yelp.com/dataset}} dataset (\text{Yelp-R}) and reviews of electronics products from Amazon Product Reviews dataset\footnote{\url{http://jmcauley.ucsd.edu/data/amazon/}} (\text{Amazon-E}) (see Table \ref{tab:labeled_data_stats}). 
    
%
%
\paragraph{Named Entity Recognition (NER):} NER is the task of identifying and categorizing named entities from unstructured text into pre-defined categories such as Person (PER), Location (LOC), Organization (ORG) etc. Figure \ref{fig: seqlabel_examples} contains an example of this task. CONLL-2003 \cite{CONLL2003} and CONLL-2012 (OntoNotes v5.0) \cite{pradhan-etal-CONLL-st-2012-ontonotes} are the commonly used \emph{labeled} datasets to build and evaluate performance for Named Entity Recognition models \cite[inter-alia]{lample-etal-2016-neural,Ma2016EndtoendSL,akbik-etal-2018-contextual}. 
We focus on the English parts of these datasets. 
CONLL-2003 contains annotations for Reuters news for 4 entity types (Person, Location, Organization, and Miscellaneous). CONLL-2012 dataset contains 18 entity types, consisting of various genres (weblogs, news, talk shows, etc.) with newswire being majority. 
We use the provided train, validation and test splits for these datasets. As newswire is the predominant genre in these datasets, as we use stories from the CNN and Daily Mail datasets\footnote{\url{https://github.com/abisee/cnn-dailymail}} (\textbf{CNN-DM}) as an unlabeled dataset from the news genre (see Table \ref{tab:labeled_data_stats}). 
	
\paragraph{Slot-labeling:} Slot-labeling is a key component of Natural Language Understanding (NLU) in dialogue systems, which involves labeling words of an utterance with pre-defined attributes - slots. For this task, we use the widely-used MIT-Movie dataset\footnote{\url{https://groups.csail.mit.edu/sls/downloads/movie/}} as labeled data which contains queries related to movie information, with 12 slot labels such as Plot, Actor, Director, etc.. An example from this dataset is demonstrated in Figure \ref{fig: seqlabel_examples}. We use the default train-test split, and create a validation set by randomly selecting 10\% of the training samples. IMDb Movie review dataset (\textbf{IMDb}) is used as in-domain unlabeled data \cite{imdbdataset} (see Table \ref{tab:labeled_data_stats}). 

\section{Models and Experimental Setup}\label{sec:modelsExp}
\begin{figure}[tp]
  \includegraphics[width=0.5\textwidth]{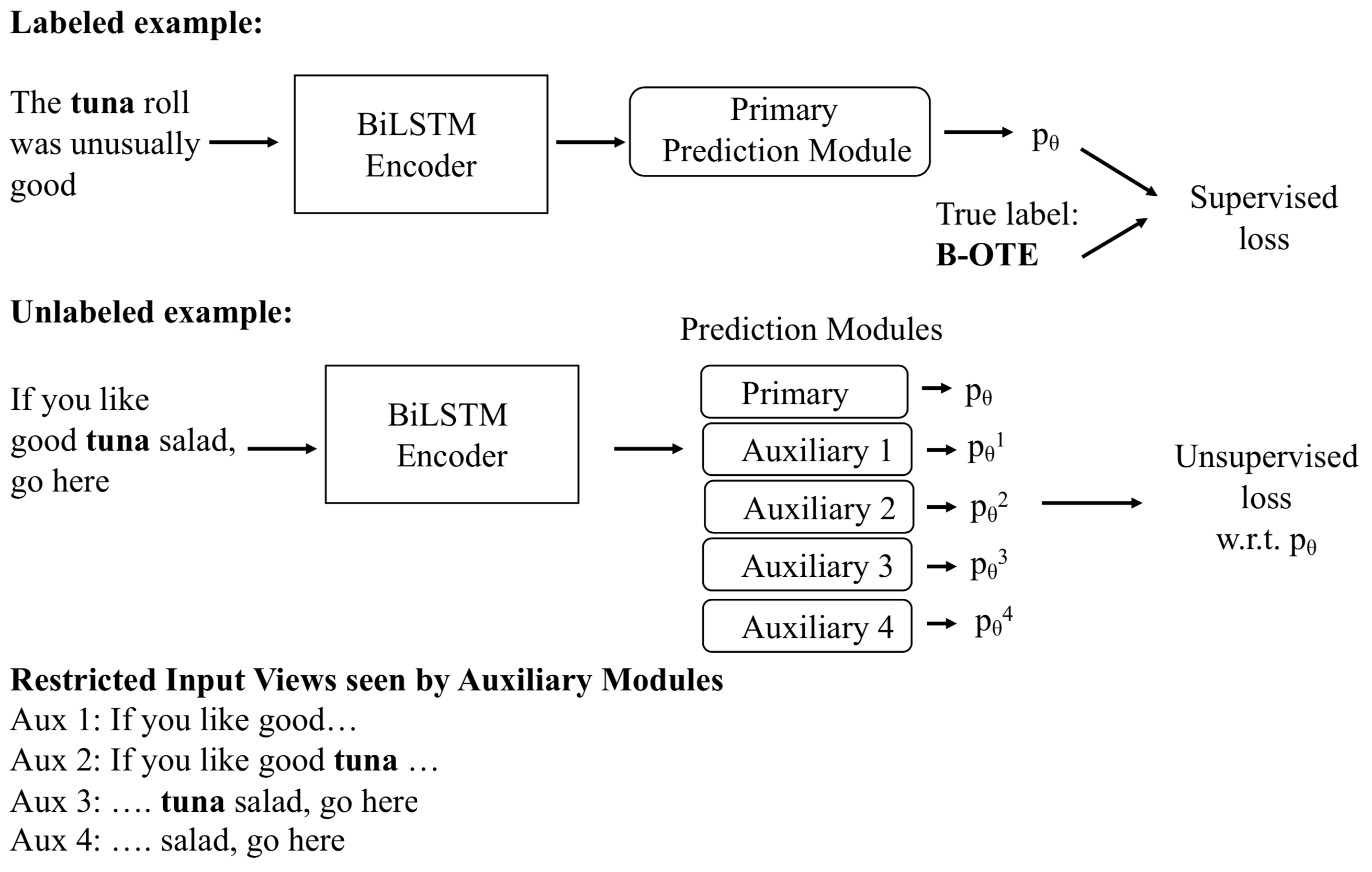}
  \caption{CVT explained using the OTE task (figure adapted from \cite{clark2018semi}). In the labeled example, \emph{tuna roll} is the OTE, hence \emph{tuna} has \textbf{B-OTE} as the gold label.}
  \label{fig: CVT_diag}
\end{figure}
\begin{table}[!ht]
\scalebox{0.8}{	
	\begin{subtable}	[t]{\textwidth}
	\textbf{Results for SemEval2016 Restaurants Dataset} \\ \\
		\begin{tabular}{|c|c|c|} \hline
			\textbf{Model} & \textbf{Unlabeled Data} & \textbf{Mean F1} \\ \hline \hline
			CVT & Yelp-R ($\sim$25.6M) & 80.08$\pm$0.18 \\ \hline
	
			BERT\textsubscript{Base} & Wiki+Books ($\sim$192M) & 75.04$\pm$1.00 \\ \hline
	
			Pre-BERT\textsubscript{Base} & Yelp-R ($\sim$261M) & 79.82$\pm$0.22 \\ \hline
	
			APBERT\textsubscript{Base} & Yelp-R ($\sim$246M) & \textbf{80.28$\pm$ 0.29} \\ \hline
	
    		DE-CNN & Yelp-R (-) & 74.37 \\ \hline

		\end{tabular}
		\label{tab: semeval16r}
	\end{subtable}
	}
 	\vspace{0.2cm}
	
	\scalebox{0.8}{
	\begin{subtable}	[t]{\textwidth} 
	    \textbf{Results for SemEval2014 Laptops Dataset} \\ \\
		\begin{tabular}{|c|c|c|} \hline
			\textbf{Model} & \textbf{Unlabeled Data} & \textbf{Mean F1} \\ \hline \hline
			
   			CVT & Amazon-E ($\sim$25.6M) & 81.77$\pm$0.24\\ \hline
   			
   			BERT\textsubscript{Base} & Wiki+Books($\sim$192M) & 80.69$\pm$0.51 \\ \hline
   			
   			Pre-BERT\textsubscript{Base} & Amazon-E ($\sim$261M) & 83.98$\pm$0.42 \\ \hline
   			
   			APBERT\textsubscript{Base} & Amazon-E ($\sim$238M) & \textbf{84.46$\pm$0.9} \\ \hline
   			
   			DE-CNN & Amazon-L (-) & 81.59 \\ \hline    
		\end{tabular}
	
		\label{tab: semeval14l}
	\end{subtable}
	}
	\caption{Model performance for OTE detection task. The \emph{same} unlabeled dataset is used for training CVT, Pre-BERT\textsubscript{Base} and APBERT\textsubscript{Base}, and \textbf{Unlabeled Data} indicates the approximate number of sentences seen by each model during training, until convergence criteria is met. \textbf{Wiki+Books} and \textbf{Amazon-L} refer to English cased Wikipedia and Books Corpus, and Amazon Laptop Reviews, respectively. \newcite{xu-etal-2018-double} propose DE-CNN, the SOTA baseline for the task. They do not specify the sizes of the unlabeled data used.}
	\label{tab: semeval}
\end{table}
We describe the various models we compare in this work and the experimental setup for each of them. Experiments are geared towards comparing the performance accuracy of the models, while also measuring impact on the environment and the resources required for training these models. Details on model architecture and training are provided in Appendix~\ref{sec:appendix}.

%
\paragraph{Cross-View Training (CVT)}
CVT is a semi-supervised approach proposed by \newcite{clark2018semi} that leverages unlabeled data in a task-specific manner. The underlying model is a two-layer CNN-BiLSTM sentence encoder followed by a linear layer and a softmax per prediction module. There are two kinds of prediction modules - \emph{primary} and \emph{auxiliary}. CVT alternates between learning from labeled and unlabeled data during training. The key idea is that the primary prediction module, which has an unrestricted view of the data, trains on the task using labeled examples, and makes task-specific predictions on unlabeled data. The auxiliary modules, with different restricted views of the unlabeled data, attempt to mimic the predictions of the primary module. Standard cross-entropy loss is minimized when learning from labeled examples, while for unlabeled examples, KL-Divergence \cite{kullback1951} between the predicted primary and auxiliary probability distributions is minimized (see  \cite{clark2018semi} for more details). We demonstrate the training strategy in Figure \ref{fig: CVT_diag}.
Thus, the model is trained to produce consistent results despite seeing partial views of the input - thereby improving underlying representations.


We use Glove 840B.300d embeddings \cite{pennington2014glove} instead of Glove 6B.300d embeddings used by the authors for a larger vocabulary coverage. For each of the labeled datasets (Section \ref{sec:taskdata}), we use the corresponding domain/task-relevant unlabeled data to train a sequence tagging model for 400K steps, with early stopping enabled using validation set convergence.
\paragraph{BERT\textsubscript{Base}} BERT \cite{devlin2019bert} has achieved state-of-the-art results on many NLP tasks. The key innovation lies in the use of bi-directional Transformers as well as the Masked Language Model (MLM) and Next Sentence Prediction (NSP) objectives used during training. Learning happens in two steps: 1) training the model on a very large generic dataset (using the two objectives above); 2) fine-tuning the learned representations on a downstream task in a supervised fashion. For our experiments we use BERT\textsubscript{Base}, which has 12 layers, 768 hidden dimensions per token and 12 attention heads and is pre-trained on the cased English Wikipedia and Books Corpus data (\textbf{Wiki+Books}). 
In order to fine-tune on the downstream sequence tagging task, the model we use consists of the BERT\textsubscript{Base} encoder, followed by a dropout layer and a classification layer that classifies each token into {B-\emph{label}, I-\emph{label}, O}, where $label \in \{label_i, label_{i+1}, ..., label_n\}$. Cross-entropy loss is the loss function used.

\paragraph{Pretrained BERT\textsubscript{Base} (Pre-BERT\textsubscript{Base})} In this setup, we use BERT\textsubscript{Base} architecture and pre-train it from scratch on the domain/task relevant unlabeled data. Each training step trains on a batch of size 256. A validation set is created from each unlabeled dataset by random sampling (details in Appendix \ref{sec:appendix}). The convergence criteria is set to be validation MLM accuracy improvement $\geq$ 0.05 when evaluated every 30K steps. We then perform the second step of fine-tuning on the downstream task data, as in the regular BERT setup. 

\paragraph{Adaptively Pretrained BERT\textsubscript{Base}(APBERT\textsubscript{Base})} Here, we start with BERT\textsubscript{Base} trained on the generic unlabeled dataset (English Wikipedia and Book Corpus) and continue pretraining on the corresponding domain/task-relevant unlabeled data (Section \ref{sec:taskdata}). Inspired by the nomenclature in \cite{Gururangan2020DontSP}, we refer to this model as Adaptively Pretrained BERT\textsubscript{Base}. Further, we perform fine-tuning on the downstream task data, as with the previous BERT models.
\begin{table}
    \scalebox{0.8}{
	\begin{subtable}[t]{\textwidth}
		\textbf{Results on CONLL-2003 dataset}\\
		\\
		\begin{tabular}{|c|c|c|} \hline
			\textbf{Model} & \textbf{Unlabeled Data} & \textbf{Mean F1} \\ \hline \hline
			CVT & CNN-DM ($\sim$17M) & 92.26$\pm$0.11 \\ \hline
		
			BERT\textsubscript{Base} & Wiki+Books ($\sim$192M) & 91.22$\pm$0.21 \\ \hline
		
			Pre-BERT\textsubscript{Base} & CNN-DM ($\sim$146M) & 85.54$\pm$0.19\\ \hline
		
			APBERT\textsubscript{Base} & CNN-DM ($\sim$138M) & 88.02$\pm$0.18 \\ \hline
		
			Cloze & Wiki+Books ($\sim$192M) & \textbf{93.5} \\ \hline
		\end{tabular}
		\label{tab: conll2003 results}
	\end{subtable}
	}
	
	
	\scalebox{0.8}{
	\begin{subtable}[t]{\textwidth}
	\vspace{0.4cm}
	\textbf{Results on CONLL-2012 dataset}\\
	\\
		\begin{tabular}{|c|c|c|} \hline
			\textbf{Model} & \textbf{Unlabeled Data} & \textbf{Mean F1} \\ \hline \hline
			CVT & CNN-DM ($\sim$18M) & 89.26$\pm$0.1 \\ \hline
			
			BERT\textsubscript{Base} & Wiki+Books ($\sim$192M) & 89.0$\pm$0.23 \\ \hline
			
    		Pre-BERT\textsubscript{Base} & CNN-DM ($\sim$146M) & 84.20$\pm$0.19 \\ \hline
    
    	    APBERT\textsubscript{Base} & CNN-DM ($\sim$138M) & 85.88$\pm$0.17 \\ \hline    
    		BERT-MRC+DSC & Wiki+Books ($\sim$192M) & \textbf{92.07} \\ \hline
		\end{tabular}
	\end{subtable} 
	}
	
	\label{tab:conll2012 results}

\caption{Model performance for NER. The \emph{same} unlabeled dataset is used for training CVT, Pre-BERT\textsubscript{Base} and APBERT\textsubscript{Base}, and \textbf{Unlabeled Data} indicates the approximate number of sentences seen by each model during training, until convergence criteria is met. Cloze \cite{Baevski2019ClozedrivenPO} and BERT-MRC+DSC \cite{Li2019DiceLF} are SOTA baselines for CONLL-2003 and CONLL-2012, respectively, for this task. \newcite{Baevski2019ClozedrivenPO} also use subsampled Common Crawl and News Crawl datasets but do not provide exact splits for these.}
\label{tab: conll}
\end{table}




\section{Results} \label{sec: Results}
We present here metrics-based and resource-based comparison of CVT and BERT models on all tasks. State-of-the-art (SOTA) baselines are included for reference.

\paragraph{Performance Metrics} We report mean F1 (with standard deviation) on the labeled test splits for each task over 5 randomized runs, and compare the models using statistical significance tests over these runs.
Further, we report the approximate number of unlabeled sentences seen by each model. Table \ref{tab: semeval} shows the results for the OTE detection task. Here, out of the 3 BERT-based variations, the best result is achieved by the APBERT\textsubscript{Base} model across both SemEval datasets. 
For SemEval2016 Restaurants, we find the mean F1 from the APBERT\textsubscript{Base} model to be comparable to that of CVT (p-value 0.26). Both models outperform the SOTA baseline.
For SemEval2014 Laptops, APBERT\textsubscript{Base} is found to have a statistically significant (p-value 0.04) higher F1 than CVT, and both models outperform SOTA. 

In Tables \ref{tab: conll} and \ref{tab: mitmovie}, we present F1 results on NER and Slot-labeling task, respectively. For all 3 datasets, we find CVT to outperform all BERT models (statistically significant for CONLL-2003 and MIT Movies dataset, at p-values 0.0086 and 0.0085, respectively). For these tasks, BERT\textsubscript{base} outperforms APBERT\textsubscript{Base} models. Furthermore, CVT outperforms SOTA for Slot-labeling task.
\begin{table}[!ht]
\centering
\scalebox{0.8}{

	\begin{tabular}{|c|c|c|} 
	\hline
	\textbf{Model} & \textbf{Unlabeled Data} & \textbf{Mean F1} \\
	\hline \hline
	CVT & IMDb ($\sim$24.1M) & \textbf{88.16$\pm$0.12} \\ \hline
	
	BERT\textsubscript{Base} & Wiki+Books ($\sim$192M) & 86.91$\pm$0.36 \\ \hline
	
    Pre-BERT\textsubscript{Base} & IMDb ($\sim$30.7M) & 85.77$\pm${0.57} \\ \hline
    
    APBERT\textsubscript{Base} & IMDb ($\sim$30M) & 86.78$\pm$0.1 \\ \hline
    
    HSCRF + softdict & - & 87.41 \\
    \hline
	\end{tabular}
	}
	\caption{Model performance for Slot-labeling. The \emph{same} unlabeled dataset is used for training CVT, Pre-BERT\textsubscript{Base} and APBERT\textsubscript{Base}, and \textbf{Unlabeled Data} indicates the approximate number of sentences seen by each model during training, until convergence criteria is met. HSCRF + softdict \cite{slotlabelbaseline} is the SOTA baseline for this task.} 
	\label{tab: mitmovie}
\end{table}
\vspace{2mm}
These results show that the CVT model, using unlabeled data in a task-specific manner, is more robust across different tasks and types of unlabeled data. For OTE detection, the unlabeled data is closely related to both domain and task, while for NER and Slot-labeling, the unlabeled data is related to genre (newswire) and domain (movies), but not necessarily to the specific tasks. In line with the findings of \newcite{Gururangan2020DontSP}, Adaptive Pre-training shows best results when using unlabeled data that is domain and task relevant (superior results for the OTE task). It is also worth noting that CVT requires significantly smaller amount of unlabeled data than the BERT models (Tables \ref{tab: semeval}, \ref{tab: conll} and \ref{tab: mitmovie}).
\begin{table}[!ht]
\centering
    \scalebox{0.82}{

\begin{tabular}{|c|c|c|c|c|c|}
    \hline
    
    \textbf{Model} & \textbf{HW} & \textbf{Hours} & \textbf{Cost} & \textbf{Power} & \textbf{CO\textsubscript{2}} \\\hline 
         CVT & 1/8 & 56 & 172 & 14.82 & 14.14 \\\hline
         Pre-BERT\textsubscript{Base} & 8/64 & 85 & 2081 & 273.62 & 261.04 \\ \hline
         APBERT\textsubscript{Base} & 8/64 & 80 & 1958 & 260.63 & 248.64 \\\hline
    \end{tabular}
    }
    \caption{Estimated CO\textsubscript{2} emissions and computational cost for CVT and BERT models, using models trained on Yelp Restaurants (Yelp-R) as an example. These computations hold for other tasks and datasets discussed in this work. \emph{HW} (hardware) refers to \#GPUs/\#CPUs used. \emph{Cost} refers to approximate cost in USD. \emph{Power} stands for total power consumption (in kWh) as combined GPU, CPU and DRAM consumption, multiplied by Power Usage Effectiveness (PUE) coefficient to account for additional energy needed for infrastructure support \cite{strubell-etal-2019-energy}. \emph{CO\textsubscript{2}} represents CO\textsubscript{2} emissions in pounds.} 
    \label{tab: resource comparison}
\end{table}

\paragraph{Resource Cost}
Table \ref{tab: resource comparison} shows computational cost and environmental impact by means of estimated CO\textsubscript{2} emissions occurring during training. We use the procedure described by \newcite{strubell-etal-2019-energy}. Tesla V100 GPUs are used for training. For computational cost, we refer to the average cost per hour for the training instances used.\footnote{\url{https://aws.amazon.com/ec2/instance-types/p3/}} To compute energy consumed, we query the NVIDIA System Management Interface \footnote{\url{https://web.archive.org/web/20190504134329/https:/developer.nvidia.com/nvidia-system-management-interface}} multiple times during training, to note the average GPU power consumption. For CPU and DRAM power usage, we use Linux's turbostat package.\footnote{\url{http://manpages.ubuntu.com/manpages/xenial/man8/turbostat.8.html}}
The models trained using Yelp Restaurants unlabeled data are used as an example in Table \ref{tab: resource comparison}, but the same computations hold for other models. Note that we do not perform initial pretraining of BERT\textsubscript{base} nor pretrain the Glove 840B.300d embeddings used in CVT, but these come at a one-time cost that we consider constant. Worth noting though, that BERT\textsubscript{base} pretraining is more expensive than Glove pretraining. As is evident, training the CVT model incurs much less financial cost than the corresponding BERT models ($\sim$11x lower than APBERT\textsubscript{base}), while also emitting lesser CO\textsubscript{2} emissions ($\sim$18x lower than APBERT\textsubscript{base}).\footnote{ If we consider just fine-tuning  BERT\textsubscript{Base} on the supervised data of the downstream task (OTE detection on SemEval2016 Restaurants Data) the numbers corresponding to Table \ref{tab: resource comparison} are: HW: 1/8, Hours: 0.283, Cost: 0.87, Power: 0.094,  CO\textsubscript{2}: 0.09. Although fine-tuning BERT on the downstream task (using supervised data) is relatively cheap, and one could amortize the cost of pre-training BERT over a large number of such tasks, this requires an understanding of what the number and type of such tasks are.} 


\section{Conclusion \& Future Work}
We compare the task-specific semi-supervised method, CVT, with a task-agnostic semi-supervised approach, BERT (with and without adaptive pre-training), on a variety of problems that can be modeled as sequence tagging tasks. 
We find that the CVT-based approach is more robust than BERT-based models across tasks and types of unsupervised data available to them. Furthermore, the financial and environmental costs incurred are also significantly lower using CVT as compared to BERT.

As part of future work, we will explore CVT on other sequence-labeling tasks such as chunking, elementary discourse unit segmentation and argumentative discourse unit segmentation, thus moving beyond entity-level spans. Moreover, other supervised tasks such as classification could also be studied in this context. Furthermore, we intend to implement CVT as a training strategy over Transformers (BERT) and compare it with Adaptively-Pretrained BERT.

\bibliography{main}

\appendix
\section{Appendices} \label{sec:appendix}
\subsection{Source Code and Data Preprocessing Steps}
For CVT, we use the official author-provided codebase.\footnote{\url{https://github.com/tensorflow/models/tree/master/research/cvt_text}} The unlabeled datasets are preprocessed to have one sentence per line using NLTK's sentence tokenizer \footnote{\url{https://www.nltk.org/api/nltk.tokenize.html}}, as required by the model. 
For BERT pretraining, we use GluonNLP's open-source code.\footnote{\url{https://github.com/dmlc/gluon-nlp/tree/master/scripts/bert}} For each unlabeled dataset, we create a randomly sampled validation set of about $30$K samples during these experiments. Unlabeled data is processed to be in the required format.\footnote{\url{https://github.com/dmlc/gluon-nlp/blob/master/scripts/bert/sample_text.txt}}

\subsection{Model Hyperparameters, Training Details and Validation F1}
Here, we enlist the hyperparameters used for each model, and describe the training process. conlleval is used as the evaluation metric for each of the models.\footnote{\url{https://www.clips.uantwerpen.be/conll2003/ner/bin/conlleval}}

\vspace{1mm}
\noindent\textbf{CVT:} A batch-size of 64 is used for both labeled and unlabeled data. We use character embeddings of size 50, with char CNN filter widths of [2,3,4], and 300 char CNN filters. Encoder LSTMs have sizes 1024 and 512, respectively for the 1st and 2nd layer, with a projection size of 512. Dropout of 0.5 for labeled examples and 0.8 for unlabeled examples is used. Base learning rate of 0.5 is used, with an adaptive learning rate scheme, using SGD with Momentum as the optimizer.

\vspace{2mm}
\noindent\textbf{Pretrained BERT\textsubscript{Base} (Pre-BERT\textsubscript{Base}) and Adaptive Pretraining BERT\textsubscript{Base} (APBERT\textsubscript{Base}):}
Batch-size of 256 is used during training. Number of steps for gradient accumulation is set to 4. BERTAdam is used as optimizer. Base learning rate used of 0.0001 is used, which is adaptive w.r.t. the number of steps. Maximum input sequence length is set to 512.

Steps at which Pre-BERT\textsubscript{Base} models converge are $\sim$1.02M for Yelp Restaurants and Amazon Electronics, $\sim$570K for CNN DailyMail, $\sim$119K for IMDb.
Model convergence steps for APBERT\textsubscript{Base}were $\sim$960K for Yelp-R, $\sim$930K for Amazon-E, $\sim$539K for CNN-DM, $\sim$117K for IMDb.

\vspace{2mm}
\noindent\textbf{BERT\textsubscript{Base} Sequence Tagging model:} Batch size for \emph{supervised} fine-tuning on the downstream task is 10. We perform manual hyperparameter tuning over learning rate (0.00001, 0.0001 and 0.001) and dropout (0.0 to 0.5 in steps of 0.1). Validation F1 is used to select the best set of hyper-parameters which were learning rate of 0.00001, dropout of 0.0 for NER and Slot-labeling, and 0.1 for OTE detection.

We demonstrate mean validation set F1 numbers for each task and dataset in Table \ref{tab: val set F1}.
\begin{table}
\centering
\scalebox{0.8}{
\begin{tabular}{|c|c|c|} \hline
\textbf{Task/Dataset} & \textbf{Model} & \textbf{Mean Val F1} \\ \hline
	\multirow{4}*{OTE/SE16-R} & CVT & 75.12$\pm$0.49 \\ \cline{2-3}
	& BERT\textsubscript{Base} & 78.65$\pm$0.27 \\ \cline{2-3}
	& Pre-BERT\textsubscript{Base} & 82.15$\pm$0.63 \\ \cline{2-3}
	& APBERT\textsubscript{Base} & 82.88$\pm$0.74 \\ \hline
	
	\multirow{4}*{OTE/SE14-L} & CVT & 81.58$\pm$1.56 \\ \cline{2-3}
	& BERT\textsubscript{Base} & 78.26$\pm$1.3 \\ \cline{2-3}
	& Pre-BERT\textsubscript{Base} & 78.40$\pm$0.49 \\ \cline{2-3}
	& APBERT\textsubscript{Base} & 79.75$\pm$1.13 \\ \hline
	
	\multirow{4}*{NER/CONLL2003} & CVT & 95.54$\pm$0.1 \\ \cline{2-3}
	& BERT\textsubscript{Base} & 95.71$\pm$0.04 \\ \cline{2-3}
	& Pre-BERT\textsubscript{Base} & 91.15$\pm$0.10 \\ \cline{2-3}
	& APBERT\textsubscript{Base} & 93.19$\pm$0.14 \\ \hline	
	
	\multirow{4}*{NER/CONLL2012} & CVT & 87.14$\pm$0.11 \\ \cline{2-3}
	& BERT\textsubscript{Base} & 88.23$\pm$0.08 \\ \cline{2-3}
	& Pre-BERT\textsubscript{Base} & 83.38$\pm$0.15 \\ \cline{2-3}
	& APBERT\textsubscript{Base} & 84.90$\pm$0.09 \\ \hline
	
	\multirow{4}*{Slot-labeling/MIT-M} & CVT & 88.31$\pm$0.27 \\ \cline{2-3}
	& BERT\textsubscript{Base} & 88.04$\pm$0.13 \\ \cline{2-3}
	& Pre-BERT\textsubscript{Base} & 87.42$\pm$0.10 \\ \cline{2-3}
	& APBERT\textsubscript{Base} & 93.1$\pm$0.06\\ \hline
\end{tabular}}
\caption{Validation Set Metrics for all Models}
\label{tab: val set F1}	
\end{table}

\end{document}